\documentclass[letterpaper, 10 pt, conference]{ieeeconf}

\IEEEoverridecommandlockouts                              

\usepackage{mathptmx} 
\usepackage{amsmath} 
\usepackage{amssymb}  
\usepackage{graphicx}
\usepackage{hyperref}
\usepackage{siunitx}
\usepackage{subcaption}
\usepackage{booktabs}
\usepackage{xcolor}
\usepackage{textcomp}
\usepackage{gensymb} 
\usepackage{multirow} 
\usepackage{comment}
\usepackage{verbatim}
\usepackage{caption}
\usepackage{mathtools}
\usepackage{array}
\newcolumntype{M}[1]{>{\centering\arraybackslash}m{#1}}
\DeclareMathOperator*{\minimize}{minimize}

\makeatletter
\def\endthebibliography{%
	\def\@noitemerr{\@latex@warning{Empty `thebibliography' environment}}%
	\endlist
}
\def\@citex[#1]#2{\leavevmode
	\let\@citea\@empty
	\@cite{\@for\@citeb:=#2\do
		{\@citea\def\@citea{,\penalty\@m\ }%
			\edef\@citeb{\expandafter\@firstofone\@citeb\@empty}%
			\if@filesw\immediate\write\@auxout{\string\citation{\@citeb}}\fi
			\@ifundefined{b@\@citeb}{\hbox{\reset@font\bfseries ?}%
				\G@refundefinedtrue
				\@latex@warning
				{Citation `\@citeb' on page \thepage \space undefined}}%
			{\@cite@ofmt{\csname b@\@citeb\endcsname}}}}{#1}}

\makeatletter
\def\endthebibliography{%
	\def\@noitemerr{\@latex@warning{Empty `thebibliography' environment}}%
	\endlist
}
\makeatother

\renewcommand{\phi}{\varphi} 

\title{\LARGE \bf
A caster-wheel-aware MPC-based motion planner for mobile robotics}

\author{Jon Arrizabalaga$^{1,2}$, Niels van Duijkeren$^{3}$, Markus Ryll$^{2}$ and Ralph Lange$^{3}$
	\thanks{The research leading to this work was funded by Robert Bosch GmbH.}%
	\thanks{$^{1}$School of Industrial Engineering and Management, Kungliga Tekniska H\"{o}gskolan, Sweden}
	\thanks{$^{2}$Autonomous Aerial Systems Lab, Department of Aerospace and Geodesy,  Technical University of Munich, Germany. E-mail: {\tt\small jon.arrizabalaga@tum.de} and {\tt\small markus.ryll@tum.de}
	}%
	\thanks{$^{3}$Robert Bosch GmbH, Corporate Research, 71272 Renningen, Germany. E-mail: {\tt\small niels.vanduijkeren@de.bosch.com} and {\tt\small ralph.lange@de.bosch.com}}
}

\usepackage{color}
\definecolor{orange}{rgb}{1.0, 0.4, 0.0}

\IEEEoverridecommandlockouts

\usepackage{tikz}
\usepackage{textcomp}
\usepackage{hyperref}
\usepackage{lipsum}

\newcommand\copyrighttext{%
	\footnotesize Published in IEEE International Conference on Advanced Robotics (ICAR), Slovenia, Ljubljana, December 2021.\newline
	 \textcopyright 2021 IEEE. Personal use of this material is permitted.
	Permission from IEEE must be obtained for all other uses, in any current or future
	media, including reprinting/republishing this material for advertising or promotional
	purposes, creating new collective works, for resale or redistribution to servers or
	lists, or reuse of any copyrighted component of this work in other works.}
\newcommand\copyrightnotice{%
	\begin{tikzpicture}[remember picture,overlay]
		\node[anchor=south,yshift=10pt] at (current page.south) {\fbox{\parbox{\dimexpr\textwidth-\fboxsep-\fboxrule\relax}{\copyrighttext}}};
	\end{tikzpicture}%
}

\begin{document}
\bstctlcite{IEEEexample:BSTcontrol} 

\maketitle
\copyrightnotice
\thispagestyle{empty}
\pagestyle{empty}

\begin{abstract}
Differential drive mobile robots often use one or more caster wheels for balance. Caster wheels are appreciated for their ability to turn in any direction almost on the spot, allowing the robot to do the same and thereby greatly simplifying the motion planning and control. However, in aligning the caster wheels to the intended direction of motion they produce a so-called bore torque. As a result, additional motor torque is required to move the robot, which may in some cases exceed the motor capacity or compromise the motion planner's accuracy. Instead of taking a decoupled approach, where the navigation and disturbance rejection algorithms are separated, we propose to embed the caster wheel awareness into the motion planner. To do so, we present a caster-wheel-aware term that is compatible with MPC-based control methods, leveraging the existence of caster wheels in the motion planning stage. As a proof of concept, this term is combined with a model-predictive trajectory tracking controller. Since this method requires knowledge of the caster wheel angle and rolling speed, an observer that estimates these states is also presented. The efficacy of the approach is shown in experiments on an intralogistics robot and compared against a decoupled bore-torque reduction approach and a caster-wheel agnostic controller. Moreover, the experiments show that the presented caster wheel estimator performs sufficiently well and therefore avoids the need for additional sensors.
\end{abstract}
\textbf{Video}: \url{https://youtu.be/NXXZKEZUi30}

\section{INTRODUCTION}
\noindent Freely turning caster wheels are often used in mobile robots -- in particular in combination with differential drive kinematics -- because of their constructional simplicity, their limited influence on the vehicle kinematics, and their ability to carry heavy loads \cite{shabalina2018comparative}. They can be found on both lightweight platforms (e.g., robotic vacuum cleaners) and on heavyweight platforms, such as Self-Driving Vehicles (SDVs) in intralogistics. However, they pose a particular challenge for precise motion planning and control, since depending on their alignment, high bore torques may arise, potentially causing deviations from the planned path or stalling the driven wheels \cite{torres2014study}.

Previous work has decoupled this problem from motion planning, treating it as a disturbance rejection problem \cite{torres2014study, beniak2017mobile} or applying velocity command filtering \cite{Schroeder2019_Enhanced_Motion_Control}. While both methods effectively mitigate the influence of the bore torque induced by the caster wheels, modifying the output of the motion planner may cause a deviation of the vehicle from the planned path, resulting in collisions or violations of other constraints. This motivates the need to address the caster wheel's physics in the motion planning stage already.

In light of its recent applications on fast dynamical systems and its support for nonlinear system models \cite{diehl2002real}, Nonlinear Model Predictive Control (NMPC) is a well-suited framework for motion planning of differential drive robots with caster wheels. NMPC can find the (locally) optimal control action considering a nonlinear cost function while incorporating constraints on nonlinear functions of states and inputs. Therefore, NMPC can address differential drive motions while accounting for caster wheel reaction torques already in the trajectory planning stage. In addition, it allows to explicitly consider limitations on velocities and accelerations.

Multiple research questions arise naturally, including how to model the caster-wheel-aware term and how the performance compares to the aforementioned decoupled approaches.

\begin{figure}[!t]
	\includegraphics[width=0.49\linewidth]{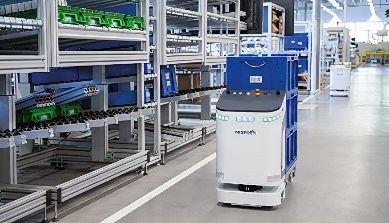}
	\includegraphics[width=0.49\linewidth]{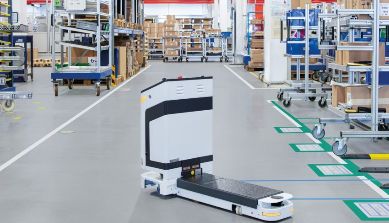}
	\caption{The ActiveShuttle by Bosch Rexroth AG is an autonomous transport system for intralogistics, used to validate the proposed caster-wheel-aware motion planner. Source: Bosch Rexroth AG.}
	\label{fig:AS_intro}
\end{figure}

 \subsection*{Contributions}
\noindent We present a caster-wheel-aware term that can be embedded in MPC-based navigation algorithms, such as trajectory tracking or path following, and is directly applicable to any mobile-robot or car-like vehicles. To the best of our knowledge, this is the first time that caster wheel awareness is accounted for in the motion planning stage. To avoid the need for additional sensors, an estimator for caster wheel states, rotation angle and rolling speed, estimator is proposed.

The key ingredients of our approach are three-fold: 1) we extend the differential drive robot model with the caster wheel orientation and rolling speed states, 2) by transforming caster wheel angular misalignment into a rolling speed offset, we derive a differentiable bore-torque penalty term, and 3) the presented caster wheel state estimator enables the proposed approach to be implemented without the need of additional sensors.

The empirical results for a commercial intralogistics SDV (see Fig.~\ref{fig:AS_intro}) indicate that the caster-wheel-aware term 1) covers longer distances without significantly reducing tracking performance, 2) decreases mean and maximum motor torques,  and 3) guarantees that constraints, such as actuator limits or collision avoidance, are fulfilled.

The remainder of this paper is structured as follows: Section~\ref{sec:related_work} provides more details on previous work on caster wheel modeling and motion planning. Section~\ref{sec:plant_model} presents a detailed plant model and Section \ref{sec:cw_aware_motionplanner} derives the caster-wheel-aware term, discusses the caster wheel state estimator, and exemplifies a complete NMPC formulation. Experiment setup and results are shown in Section~\ref{sec:experiments} before Section~\ref{sec:conclusion} presents the conclusions.

\section{RELATED WORK}\label{sec:related_work}

\noindent The dynamics of differential drive mobile robots are a well-studied problem and there exist a number of suitable models \cite{lavalle2006planning}. Due to the "omni-directional" nature of the caster wheels, their dynamical influence has been widely neglected when modeling mobile robots. An exception is \cite{staicu2009dynamics}, where the principle of virtual work and second order Lagrange equations are applied to model the kinematics and dynamics of a differential drive robot with a centered caster wheel located at the front side. Another interesting case is \cite{Schroeder2019_Enhanced_Motion_Control}, where Schr\"{o}der \textit{et al.}\ model a caster wheel and its respective reaction torque, also referred to as the bore torque. For this purpose, the authors propose a modified version of the tire model by Zimmer and Otter~\cite{zimmer2010real}.

The mitigation of the caster wheel bore torques for differential drive robots has been studied previously in \cite{torres2014study,beniak2017mobile, Schroeder2019_Enhanced_Motion_Control}. Firstly, in \cite{torres2014study} Torres \textit{et al.}\  model the caster wheel dynamics and limit the motor torques according to the identified threshold at which the driven wheels start spinning. This method requires knowledge of the disturbance's uncertainty and a good understanding of the edge cases, which strongly depends on many parameters, such as the overhang's geometry and surface-wheel friction, compromising its applicability to other platforms. 

In \cite{beniak2017mobile}, a kinematic model of the caster wheel is implemented alongside a feedback-feedforward controller. Similarly to the previous work, such an approach lacks generalizability and is decoupled from the motion planner, jeopardizing the optimality of the commanded velocities.

Lastly, in \cite{Schroeder2019_Enhanced_Motion_Control}  Schr\"{o}der \textit{et al.}\ study the impact of caster wheels dynamics on motion control and present a filter -- named \emph{path filter} -- on the commanded velocities. To reduce bore torques, this filter modifies the velocity commands depending on the estimated caster wheel state. They observe that the effects of high bore torques are most apparent for rotations on the spot or at low longitudinal velocities. Their path filter prevents these maneuvers by modifying the desired velocity commands sent from the motion planner to the motor controller. The authors report a decrease of $71\%$ of the maximum motor torque in a simulation model of a predecessor of the ActiveShuttle while driving at $0.1\,\mathrm{m/s}$ and carrying a payload of $150\,\mathrm{kg}$. However, the path filter is not integrated into the motion planner and, consequently, modifies the velocity commands without taking other constraints into account. This may lead to collisions in tight areas, such as parking or driving through narrow corridors. Thus, the path filter succeeds in reducing the caster wheel reaction torques, but at the cost of degrading navigation performance.

In contrast, our proposed approach incorporates the caster wheel physics directly in motion planning and is therefore able to account for the aforementioned constraints.
\section{MODELLING OF THE ROBOT}\label{sec:plant_model}
\noindent A differential drive robot with caster wheels consists of two independent subsystems: the differential drive kinematics and the caster wheel kinematics.
\subsection*{Notation}
\begin{figure}[t]
  \centering
	\includegraphics[width=0.65\linewidth]{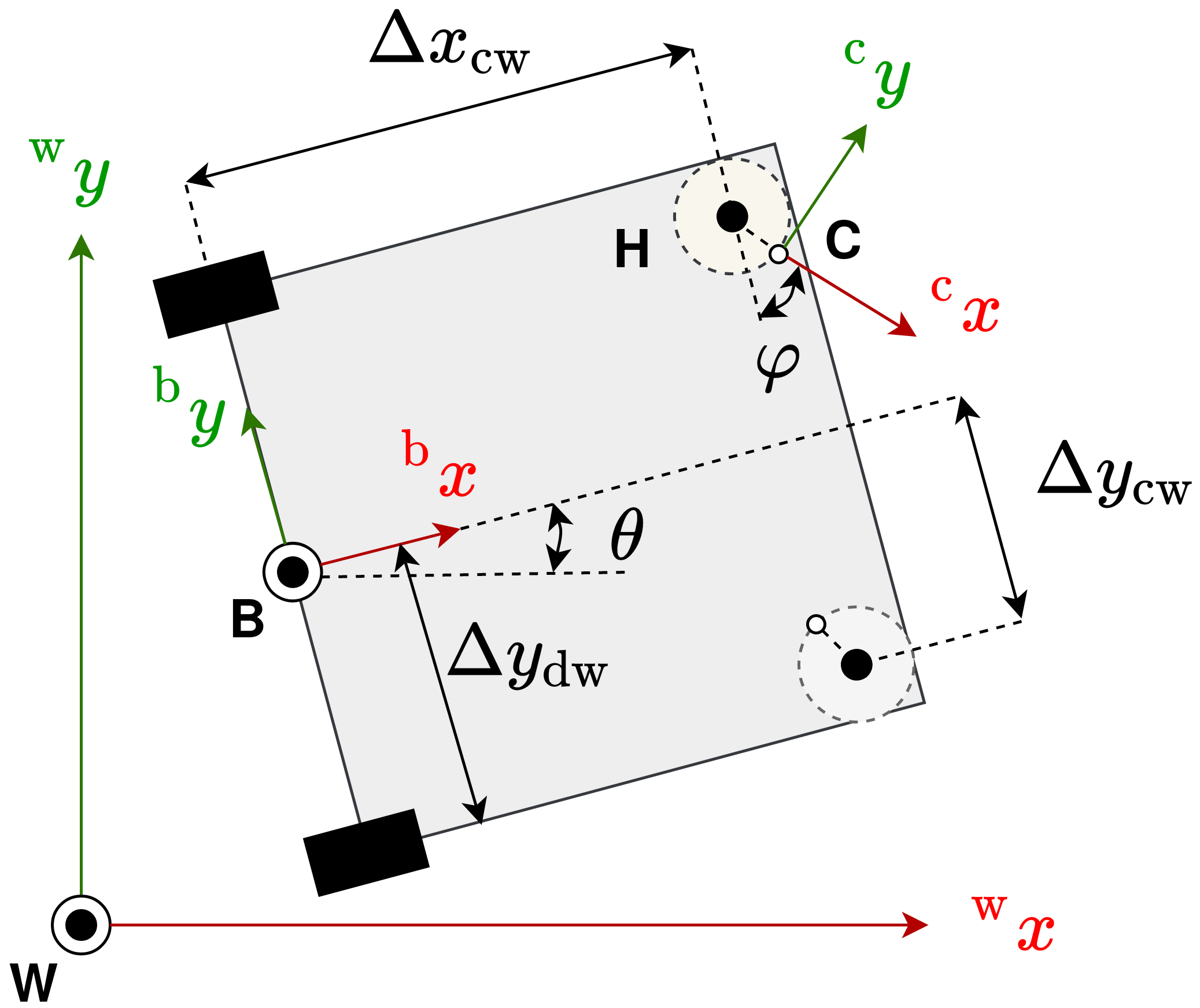}
	\caption{Top view of a differential drive with two symmetric caster wheels. $\Delta x_{\text{cw}}$, $\Delta y_{\text{cw}}$, $\Delta y_{\text{dw}}$ represent distances from the center-point $B$ to the caster and driven wheels.}
	\label{fig:diagram_robot}
\end{figure}

As depicted in Fig.~\ref{fig:diagram_robot}, we choose a world reference frame $\{\text{w}\}$ in point $\text{W}$, a body-fixed frame $\{\text{b}\}$ attached to the center point $\text{B}$ and a caster wheel-fixed frame $\{\text{c}\}$ located in the rolling center of the caster wheel $\text{C}$. Point $\text{H}$ refers to the hinge between the caster wheel and the bodywork. Regarding nomenclature,  ${}^\text{d}_\text{q}(\cdot)_\text{P}\;$ denotes the quantity $(\cdot)$ for point $\text{P}$, with respect to frame $\{\text{q}\}\,$, and expressed in frame $\{\text{d}\}$. If the subscript is omitted, it is considered to be equal to the superscript. The estimate of $(\cdot)\;$ is given by $\hat{(\cdot)}$.

\subsection{Differential Drive}
\noindent The robot is modeled to move in plane, as illustrated in Fig.~\ref{fig:diagram_robot}.
Taking longitudinal and rotational velocities ($v$ and $\omega$) as inputs, its position and orientation in the world frame can be defined as follows: 
\begin{equation} \label{eq:diff_robot}
\frac{\mathrm{d}}{\mathrm{d}t}{}^\text{w}\vec{\text{WB}}
=
v\, \begin{bmatrix}
\cos(\theta) \\\sin(\theta)
\end{bmatrix}, \; \frac{\mathrm{d}\theta}{\mathrm{d}t} = \omega
\end{equation}

\subsection{Caster wheel}

\begin{figure}[!t]
	\centering
	\includegraphics[width=0.65\linewidth]{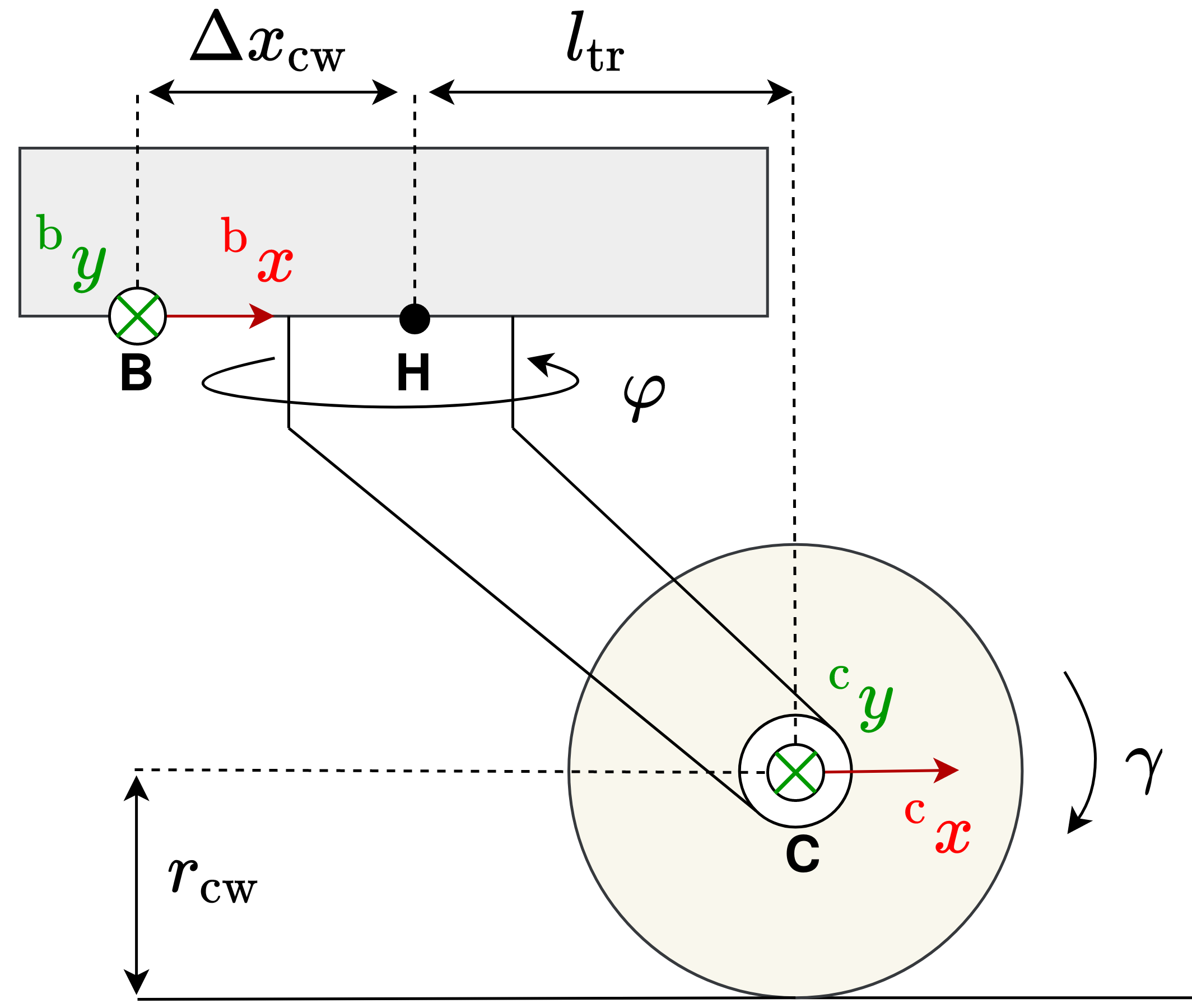}
	\caption{Side view of caster wheel diagram. Points $\text{B}$, $\text{C}$ and $\text{H}$ represent the origin of the robot, the hinge between robot-frame and overhang and the rolling center of the caster wheel.}
  \label{fig:cw}
\end{figure}

\noindent Next, we derive kinematic equations for the yaw and rolling angles ($\varphi$ and $\gamma$) of a caster wheel. Since the velocity of the origin is known, the velocity of the caster wheel rolling center in the body frame can be defined as
\begin{equation*}
^\text{b}_{\text{w}}\vec{v}_\text{C} = {}^\text{b}_\text{w}\vec{v}_\text{B}+\vec{\dot{\theta}} \times {}^\text{b}\vec{\text{BH}}+\vec{\dot{\varphi}} \times {}^\text{b}\vec{\text{HC}}\, ,
\end{equation*}
which yields
\begin{equation} \label{eq:vb1}
^\text{b}_{\text{w}}\vec{v}_\text{C} =
\begin{bmatrix}
v-\omega \, \Delta y_{\text{cw}}+\dot{\varphi}\,  \sin(\varphi)\, l_{\text{tr}} \\ \omega\, \Delta x_{\text{cw}} - \dot{\varphi}\, \cos(\varphi)\, l_{\text{tr}}
\end{bmatrix} \, ,
\end{equation}
where $l_{\text{tr}}$ is the trail of the caster wheel. Assuming there is no slip, the rolling center velocity in the caster wheel frame is
$
^\text{c}_{\text{w}}\vec{v}_\text{C} =
\begin{bmatrix}
\dot{\gamma} \, r_{\text{cw}}, 0
\end{bmatrix}
$. Rotating this velocity to the body-fixed frame results in
\begin{equation*}
^{\text{b}}_{\text{w}}{\vec{v}_\text{C}} =
\begin{bmatrix}
\cos \varphi & -\sin \varphi  \\ \sin \varphi & \cos \varphi \\
\end{bmatrix}
\cdot
\begin{bmatrix}
\dot{\gamma}\, r_{\text{cw}} \\ 0
\end{bmatrix}
=
\begin{bmatrix}
\dot{\gamma}\, r_{\text{cw}}\, \cos{\varphi} \\ \dot{\gamma}\, r_{\text{cw}}\, \sin{\varphi}
\end{bmatrix} \, .
\end{equation*}

\noindent Combining this equation with \eqref{eq:vb1}, the caster wheel rotating and rolling angles, $\varphi$ and $\gamma$, can be expressed by the following differential equations:
\begin{subequations} \label{eq:cw2}
	\begin{gather} 
	\dot{\varphi} = -\frac{1}{l_{\text{tr}}}\left[(v-\omega\, \Delta y_{\text{cw}})\, \sin(\varphi) - \omega\, \Delta x_{\text{cw}}\, \cos(\varphi)\right] \label{eq:cw}\\
	\dot{\gamma} = \frac{1}{r_{\text{cw}}}\left[(v-\omega\, \Delta y_{\text{cw}})\, \cos(\varphi) + \omega\, \Delta x_{\text{cw}}\, \sin(\varphi)\right]\label{eq:gamma}
	\end{gather}
\end{subequations}

\noindent The analytical expressions of the respective steady state angles ($\varphi_\text{ss},\dot{\gamma}_\text{ss}$) for given velocity commands $v$, $\omega$ are useful for anticipating undesired caster wheel motions that could result in increased bore torques. Taking the derivative of \eqref{eq:cw} and solving the equation for $\varphi_{\text{ss}}$ when the derivative is $0$ leads to
\begin{equation}\label{eq:phiss}
\varphi_{\text{ss}} = \text{atan}\left(\frac{\omega\, \Delta x_{\text{cw}}}{v-\omega\, \Delta y_{\text{cw}}}\right) \, .
\end{equation}
According to this expression, for given velocity states, the caster wheel can converge to two different angles separated by $\pi \,\text{rad}$, i.e., the caster wheel can either align forward or backwards. Another way to look at this is by  defining the error function $e = \varphi - \varphi_{ss}$ and looking into its equilibrium points,  $\bar{e} = \{0,\pi\}$. The respective eigenvalues obtained from linearization reveal that the first equilibrium is stable, while the second is not. In other words, if the caster wheel is facing backwards a small disturbance is sufficient for making it converge to the forward position. The stability of both equilibrium points can be observed in Fig.~\ref{fig:eigenvalues}, which depicts the eigenvalues computed by evaluating the linearization of \eqref{eq:cw} for a grid of velocity commands ($v$,$\omega$). From now onward, the steady state caster wheel rotation angle $\varphi_{ss}$  will exclusively refer to the stable equilibria.

\noindent Combining \eqref{eq:gamma} with the term obtained from \eqref{eq:phiss} defines the caster wheel steady state rolling speed $\dot{\gamma}_{\text{ss}}\,$:
\begin{equation} \label{eq:gammass_all}
\dot{\gamma}_{\text{ss}} = \frac{1}{r_\text{cw}}\, \left[(v-\omega\, \Delta y_{\text{cw}})\, \cos(\varphi_{\text{ss}}) + \omega\, \Delta x_{\text{cw}}\, \sin(\varphi_{\text{ss}})\right] .
\end{equation}

\noindent Using the trigonometric relationships $\sin(\text{atan}(x)) = \frac{x}{\sqrt{1+x^2}}$ and $\cos(\text{atan}(x)) = \frac{1}{\sqrt{1+x^2}}$, \eqref{eq:gammass_all} can be simplified to
\begin{equation} \label{eq:gammadot_ss}
\dot{\gamma}_{\text{ss}} = \frac{1}{r_\text{cw}}\,\sqrt{(v-\omega\, \Delta y_{\text{cw}})^2+(\omega\, \Delta x_{\text{cw}})^2} \: .
\end{equation}

\subsection{Plant model}
\noindent We model the ActiveShuttle as a differential drive with two caster wheels at the front axle (see Fig.~\ref{fig:diagram_robot}). Notice that the equations of motion have been derived for an arbitrary caster wheel, and therefore, they can also be used for modeling other setups, such as mobile robots with a single-centered or multiple-distributed caster wheels. To ensure smoothness of the planned motions, the kinematics chain of the differential drive model in \eqref{eq:diff_robot} is extended to the accelerations domain, shifting the input states from velocities ($v$, $\omega$) to accelerations ($a$, $\alpha$). By appending the additional state equations \eqref{eq:cw} for both caster wheels to the extended differential drive model results in a plant model with seven states:

\begin{subequations} \label{eq:plant}
	\begin{align} 
	\dot{x}_1 &= x_{4}\,\cos(x_{3}) \\
	\dot{x}_2 &= x_{4}\,\sin(x_{3})\\
	\dot{x}_3 &= x_{5}\\
	\dot{x}_4 &=  u_1\\
	\dot{x}_5 &=  u_2\\
	\dot{x}_6 &= -\frac{1}{l_{\text{tr}}}[(x_4-x_5 \Delta y_{\text{cw}}) \sin(x_6) - x_5 \Delta x_{\text{cw}} \cos(x_6)]\\
	\dot{x}_7 &= -\frac{1}{l_{\text{tr}}}[(x_4+x_5\Delta y_{\text{cw}})\sin(x_7)- x_5\Delta x_{\text{cw}}\cos(x_7)] \text{,}
	\end{align}
\end{subequations}
where $\vec{x} = \begin{bmatrix}
	x_1, \ldots, x_7 
	\end{bmatrix} = \begin{bmatrix}
\text{WB}_x, \text{WB}_y, \theta, v, \omega,\varphi_{\text{L}}, \varphi_\text{R} 
\end{bmatrix}$ and $\vec{u}= \begin{bmatrix}u_1, u_2 \end{bmatrix} = \begin{bmatrix}a, \alpha \end{bmatrix}$.  

\section{CASTER-WHEEL-AWARE MOTION PLANNING} \label{sec:cw_aware_motionplanner}
\noindent We are interested in leveraging the influence of caster wheels when finding the optimal commands that trade off between tracking a predefined trajectory and minimizing control effort. When doing so, we also require that other constraints, such as hardware limitations or collision avoidance, are fulfilled. To this end, we take a modular approach, where we present a \emph{caster-wheel-aware-term} $J_{\text{cw}}$ that can be embedded into the cost function of an MPC-based navigation algorithm: 
\begin{equation}\label{eq:cost_function}
	J = J_{\text{navigation}}+J_{\text{cw}}.
\end{equation}

\subsection{Caster-wheel-aware term}
\noindent To fulfill the characteristics listed above, the caster-wheel-aware term, $J_\text{cw}$, needs to 1) penalize undesired caster wheel motions, 2) be two times differentiable and 3) its derivatives need to be defined and continuous in the entire input range.

From \cite{Schroeder2019_Enhanced_Motion_Control}, caster wheel bore torques are highest when rolling at slow speeds and misaligned with respect to the robot heading. Therefore, the caster-wheel-aware term can either be tackled through rotation angles or rolling speeds. For both cases, penalizing differences in the caster wheel's \emph{current} and \emph{future} states mitigates abrupt changes in their motion, diminishing the presence of bore torques. The \emph{current} state is assumed to be measured or estimated, while the \emph{future} state represents the steady state for given commands.

Taking the first approach, penalizing the difference in rotation angles leads to $\Delta \varphi = \varphi_{\text{ss}} - \varphi$. Since the $\text{atan}$ in \eqref{eq:phiss} cannot distinguish between all quadrants, we replace it by its 2-argument version, resulting in $\varphi_{\text{ss}} = \text{atan2}\left(\omega\, \Delta x_{\text{cw}},{v\pm \omega\, \Delta y_{\text{cw}}}\right)$. However, $\text{atan2}$ is discontinuous in the origin for $v-\omega\, \Delta y_{\text{cw}} < 0$, and thus, does not fulfill the third requirement and can therefore cause numerical problems in the optimization algorithm. 

A solution can be found in the second alternative, where differences in rolling speeds are penalized  $\Delta \dot{\gamma} = \dot{\gamma}_{\text{ss}} - \dot{\gamma}$. Since in the derivation of the expression for $\dot{\gamma}_{\text{ss}}$ in \eqref{eq:gammadot_ss}, atan2 is replaced by a square root, its first two derivatives of the term become continuous in the entire input range. However, when the operand is 0, the derivatives of this term are not defined, making this term problematic. To accommodate for this, a tiny positive constant is added, resulting in a strictly positive operand: 

\begin{equation}\label{eq:gammass_1}
\dot{\gamma}_{\text{ss}} \approx \Gamma = \frac{1}{r_{\text{cw}}}\,\sqrt{(v \pm \omega\, \Delta y_{\text{cw}})^2+(\omega\, \Delta x_{\text{cw}})^2 + \zeta}
\end{equation}

\noindent where the small $\zeta > 0$ ensures that the square root remains differentiable when $[v,\omega]=0$. Therefore, we propose a caster-wheel-aware term that penalizes the difference between the caster wheel rolling speed, $\dot{\gamma}$ in \eqref{eq:gamma}, and its approximated steady state value, $\Gamma$ in \eqref{eq:gammass_1}, i.e., $ J_\text{cw} = \dot{\gamma}-\Gamma$.

\paragraph*{Multiple caster wheels} $\Gamma$ can be directly extended for multiple caster wheels. For example, applying it for both caster wheels in plant model \eqref{eq:plant} yields 
\begin{equation}\label{eq:gammass_2}
\begin{bmatrix}
\Gamma_{\text{L}} \\ \Gamma_{\text{R}} \\
\end{bmatrix}
=
\frac{1}{r_{\text{cw}}}\,
\begin{bmatrix}
\left[(x_4-x_5\, \Delta y_{\text{cw}})^2+(x_5\, \Delta x_{\text{cw}})^2 + \zeta\right]^\frac{1}{2} \\
\left[(x_4+x_5\, \Delta y_{\text{cw}})^2+(x_5\, \Delta x_{\text{cw}})^2 + \zeta \right]^\frac{1}{2}\\
\end{bmatrix} \, .
\end{equation}

\begin{figure}[t]
	\centering
  	\vspace{2mm} 
	\includegraphics[width=0.95\linewidth]{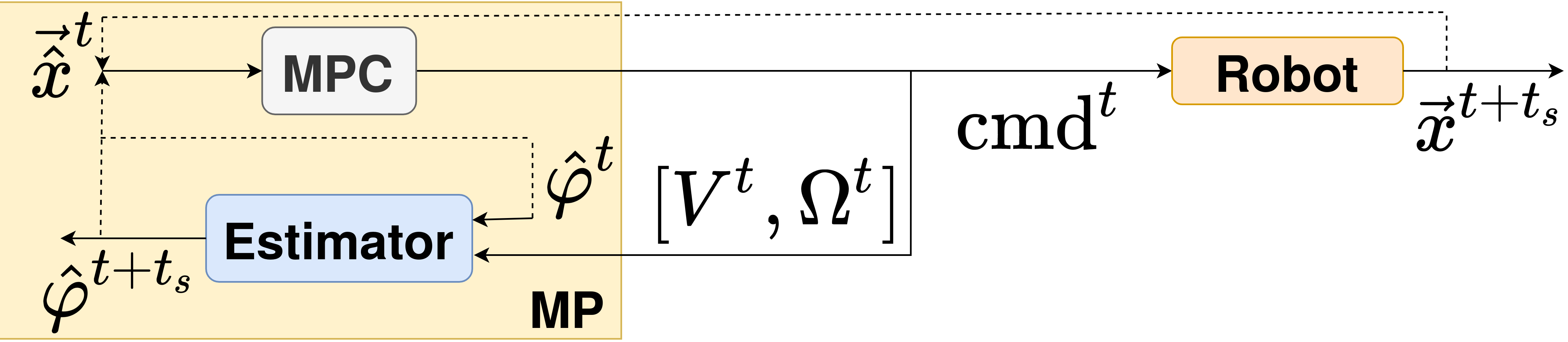}
	\caption{Block diagram of the presented caster-wheel-aware motion planner (MP, yellow box).}
  \label{4fig:estimator_layout}
\end{figure}

\subsection{ Caster wheel state estimator}
\noindent The caster-wheel-aware term $J_\text{cw}$ derived in the previous subsection requires knowing the caster wheel rolling speed. To avoid the need for additional sensors, we present an observer that estimates the full caster wheel state, rotation angles and rolling speeds ($\hat{\varphi}$, $\hat{\dot{\gamma}}$). Applying \eqref{eq:cw} to both caster wheels leads to
\begin{equation} \label{eq:observer_model}
\begin{bmatrix}
\dot{\xi}_1 \\ \dot{\xi}_2 \\
\end{bmatrix}
=
-\frac{1}{l_\text{tr}}\, 
\begin{bmatrix}
(\eta_1-\eta_2\, \Delta y_{\text{cw}})\sin(\xi_1) - \eta_2 \Delta x_{\text{cw}}\cos(\xi_1)\\ (\eta_1+\eta_2 \Delta y_{\text{cw}})\sin(\xi_2) - \eta_2\Delta x_{\text{cw}}\cos(\xi_2) \\

\end{bmatrix}\, ,
\end{equation}

\noindent where \(\vec{\xi} = \begin{bmatrix}\hat{x}_6, \hat{x}_7\end{bmatrix}\), for the states given in \eqref{eq:plant}, and \(\vec{\eta} = \begin{bmatrix}V, \Omega\end{bmatrix}\), referring to real longitudinal and angular velocities. As illustrated in Fig.~\ref{4fig:estimator_layout}, the estimator takes previous estimates ($\hat{\varphi}^{\,t}$) and real velocities ($V^t,\Omega^t$), measured from sensor data, as inputs. Analogously from \eqref{eq:gamma}, the estimated rotation angles, $\vec{\xi}$, together with the true velocities, $\vec{\eta}$, are sufficient to estimate the rolling speeds:

\begin{equation} \label{eq:observer_model_2}
	\begin{bmatrix}
		\hat{\dot{\gamma}}_\text{L} \\ \hat{\dot{\gamma}}_\text{R}
	\end{bmatrix}
	=
	\frac{1}{r_{\text{cw}}}\,
	\begin{bmatrix}
		(\eta_1-\eta_2\, \Delta y_{\text{cw}})\, \cos(\xi_1) + \eta_2\, \Delta x_{\text{cw}}\, \sin(\xi_1)\\
		(\eta_1+\eta_2\, \Delta y_{\text{cw}})\, \cos(\xi_2) + \eta_2\, \Delta x_{\text{cw}}\, \sin(\xi_2)\\
		
	\end{bmatrix}\, .
\end{equation}

Defining $\hat{e} = \hat{\varphi} - \varphi_{ss}$, it can be shown by the Lyapunov function $V(\hat{e})=\frac{1}{2}\cdot \hat{e}^2$ that the caster wheel steady state rolling angle, $\varphi_{\text{ss}}$, is asymptotically stable for any set of non-zero (static) input commands. Apart from proving that the estimates will converge to the correct values, this property can be leveraged for defining the initial caster wheel state.
\begin{figure}[t]
	\vspace{2mm} 
	\centering
	\includegraphics[width=1\linewidth]{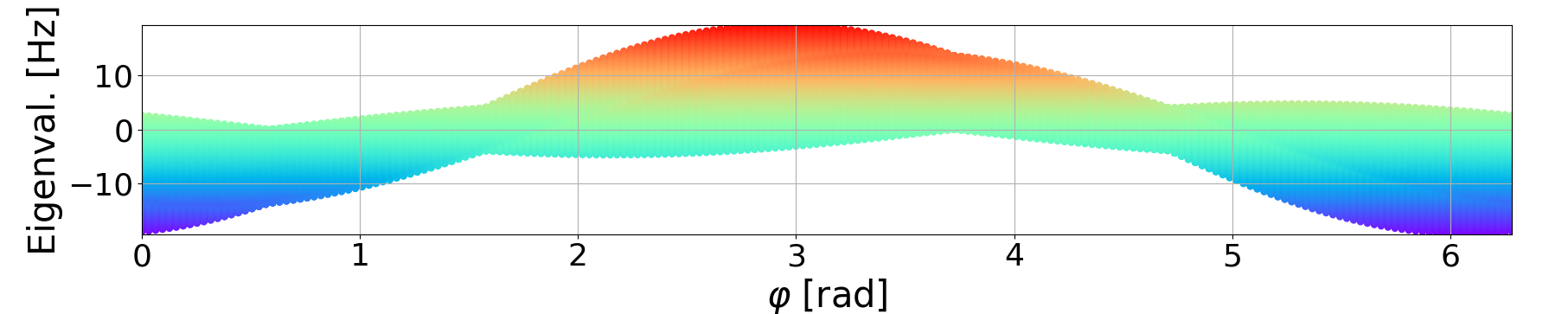}
	\caption{Eigenvalues of \eqref{eq:cw}, evaluated for a grid of velocity commands ($v$,$\omega$). Both the y-axis and the color code refer to the speed of the eigenvalues.}
	\label{fig:eigenvalues} 
\end{figure}
\subsection {Exemplary trajectory-tracking NMPC formulation}
\noindent The caster-wheel-aware term $J_\text{cw}$, can be paired with MPC-based navigation algorithms. Next, we illustrate a case-specific showcase by combining  it with the simplest form of navigation control, namely trajectory tracking. In such a framework, the predefined path is discretized in time, so that the distance from the robot to a virtual time-based pose reference is penalized, i.e., $ J_\text{nav} = \vec{p}-\vec{p}_\text{ref}^{\,t}, \, \forall t\in[0,t_\text{end}]$ and $\vec{p} = \begin{bmatrix}x_1,x_2,x_3\end{bmatrix}$. For further details please refer to \cite{kunhe2005mobile}.

To find the optimal commands of cost function \eqref{eq:cost_function} under the model \eqref{eq:plant}, while fulfilling constraints in states and inputs, the following nonlinear program (NLP) is formulated:

\begin{subequations}\label{eq:NLP}
	\begin{alignat}{2} 
	&\!\minimize_{\vec{x}_1,..., \vec{x}_{N},\vec{u}_1,...,\vec{u}_{N-1}}&& \sum_{k=1}^{N+1}
	\left|\left|\Delta p^{\,k}\right|\right|_{Q_{\text{nav}}}^2 + \left|\left| \Delta \dot{\gamma}^{\,k}\right|\right|^2_{Q_{\text{cw}}} +
	\left|\left|u^{\,k}\right|\right|_{Q_{\text{u}}}^2 \\ \nonumber
	&\text{subject to} &      & \\
	&	&	&\vec{x}^{\,k+1} = F(\vec{x}^{\,k},\vec{u}^{\,k}) \\
	&	&	&\Delta p^{\,k} = \begin{bmatrix} x_{\text{ref}}^{\,k} - x_{1}^k \\
		y_{\text{ref}}^k-x_{2}^k \\ \theta_{\text{ref}}^k-x_{3}^k
		\end{bmatrix} \\
	&	&	&\Delta \dot{\gamma}^{\,k} = \begin{bmatrix}
	\hat{\dot{\gamma}}^{\,k}_\text{L}-\Gamma_{\text{L}}^{\,k} \\
	\hat{\dot{\gamma}}^{\,k}_\text{R}-\Gamma_{\text{R}}^{\,k}
	\end{bmatrix} \\				
	&                  &      & v_{\text{max}} \geq x_4^{\,k} \geq
	v_{\text{min}} \\
	&                  &      & \omega_{\text{max}} \geq x_5^{\,k} \geq
	\omega_{\text{min}} \\
	&                  &      & a_{\text{max}} \geq
	u_1^{\,k}-u_2^{\,k}\,\frac{\Delta y_{\text{dw}}}{2} \geq
	a_{\text{min}} \\
	&                  &       & a_{\text{max}} \geq
	u_1^{\,k}+u_2^{\,k}\,\frac{\Delta y_{\text{dw}}}{2} \geq
	a_{\text{min}},
	\end{alignat}
	where the  approximated caster wheel steady state rolling speeds, $\Gamma$, are defined by \eqref{eq:gammass_2}, states $x_6^ k$, $x_7^k$ are estimated from the observer \eqref{eq:observer_model} and the caster wheel rolling speeds, $\hat{\dot{\gamma}}^{\,k}_\text{L}$ and $\hat{\dot{\gamma}}^{\,k}_\text{R}$  are given by \eqref{eq:observer_model_2}. 
\end{subequations}

\section{EXPERIMENTS}\label{sec:experiments}
\noindent To evaluate the proposed caster-wheel-aware motion planner, we conduct experiments with the ActiveShuttle (see Fig.~\ref{fig:AS_intro}). The NLP \eqref{eq:NLP} is implemented in CasADi \cite{Andersson2019} and solved by IPOPT~\cite{waechter2011introduction} with a prediction time of 2s and a sampling frequency of 20Hz, based on the maximum Eigenvalues in Fig.~\ref{fig:eigenvalues}.

Before we present the results, clarifications in nomenclature should be made. Firstly, \emph{Aware Planner} refers to the here presented caster-wheel-aware motion planner: $ J_{\text{aw}} = J_{\text{nav}} + J_{\text{cw}}$. Secondly, \emph{Agnostic Planner} does not include the caster-wheel-aware term:  $J_\text{ag} = J_{\text{nav}}$. Thirdly, \emph{Path Filter Planner} refers to the method proposed in \cite{Schroeder2019_Enhanced_Motion_Control}, where the path filter is combined with the Agnostic Planner: $\left[J_\text{ag}\right] +\left[\text{PF}\right]$. 

\subsection{Validation of estimator}
\begin{figure}[t]
	\vspace{2mm}
	\centering
	\includegraphics[width=1\linewidth]{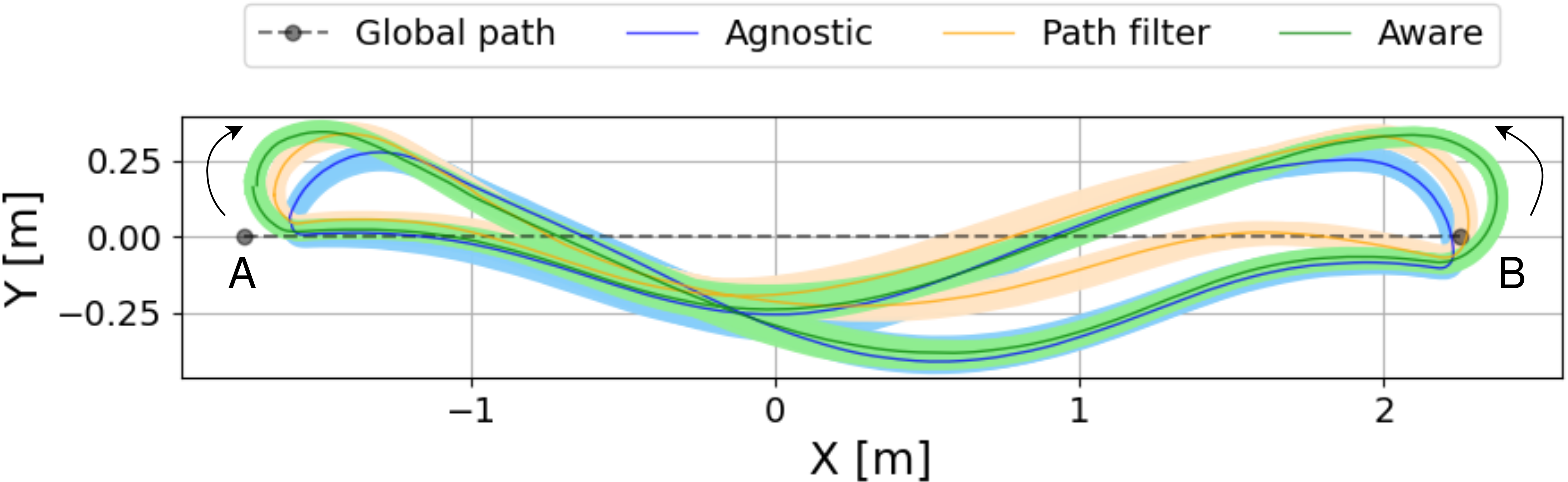}
	\caption{Trajectories  of Agnostic, Aware, and Path Filter Planner when navigating along the $4\,\mathrm{m}$ line global path. 
	\label{6fig:hardware_navigation_trajectories}}
\end{figure}

\noindent At first, we validate the caster wheel state estimator. We compare ground-truth data, obtained from placing an encoder at each caster wheel, to rotation angles observed from the estimator proposed in Section \ref{sec:cw_aware_motionplanner}. For this purpose, we make the ActiveShuttle navigate back and forth along a $4\,\mathrm{m}$ straight line, obliging it to perform a hairpin turn once it reaches the end (see Fig.~\ref{6fig:hardware_navigation_trajectories}).

\begin{figure}[t]
	\vspace{2mm} 
	\centering
	\includegraphics[width=.975\linewidth]{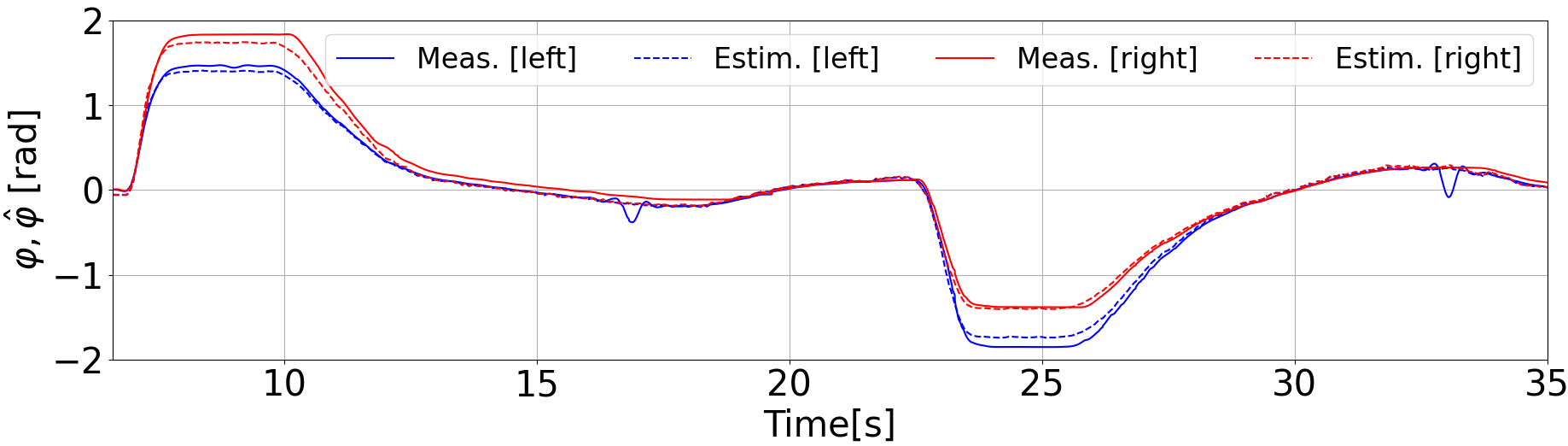} 
	\caption{Comparison of measured (continuous, $\varphi$) and estimated (dashed, $\hat{\varphi}$) caster wheel rotation angles.}
	\label{6fig:sensor_measurements}
\end{figure}

The respective measurements and estimations are given in Fig.~\ref{6fig:sensor_measurements}. Apart from the spikes in the left wheel caused by sensor failure, estimations and measurements only differ in steady state angles by a magnitude of $0.1\,\mathrm{rad}$ approximately. The respective RMSE values are $0.0292\,\mathrm{rad}$ and $0.0230\,\mathrm{rad}$ for left and right caster wheels. This shows that the observer provides valid estimates of the caster wheel angle and, consequently, no additional sensors are required for implementing the proposed caster-wheel-aware term.

\subsection{Case study I: Navigation and hairpin turns}
\noindent To observe the capacity of the Aware Planner to trade off between navigation performance and bore torques, we compare it to the Agnostic and Path Filter Planners. Since bore torques are sensitive to hairpins, the global path described above is reused. To account for uncertainty, the experiment is repeated ten times for all three planners. 

\begin{figure}[b]
	\centering
	\includegraphics[width=1\linewidth]{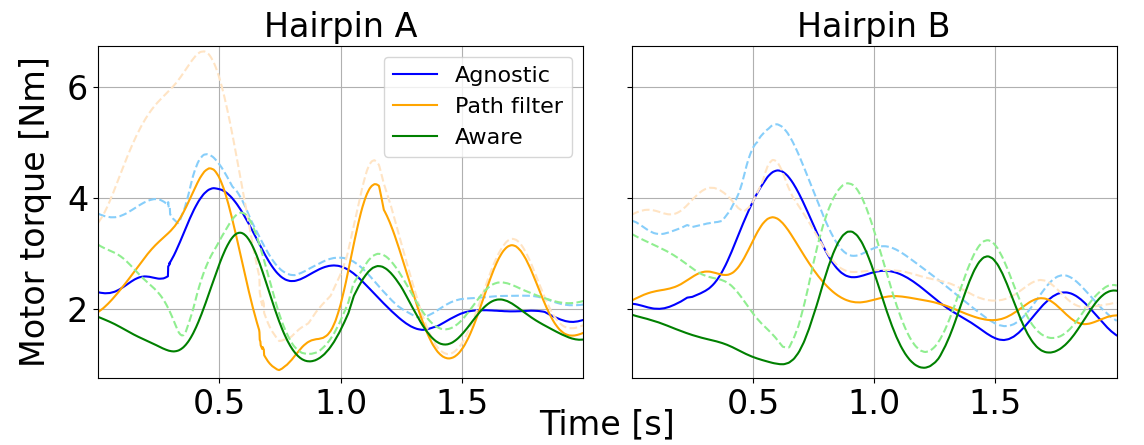}
	\caption{Comparison of mean (continuous) and maximum (dashed) motor torques when driving along hairpins.}
	\label{6fig:torques_navigation}
\end{figure}

Fig.~\ref{6fig:hardware_navigation_trajectories} depicts the resultant trajectories, where the solid lines show the average of the ten iterations, while the shaded regions illustrate the maximum deviations. 
The respective \emph{averaged-mean} and \emph{averaged-maximum} motor torques required for driving through both hairpins are given in Fig.~\ref{6fig:torques_navigation}. These terms refer to mean or maximum of both motor torques, after being averaged across the ten iterations. All the results have been quantified in Table \ref{6tab:main_table}. Notice that the RMSE refers to the area between the swept path by the robot and the reference path and hence, the lower values are better.

According to Table \ref{6tab:main_table} and Fig.~\ref{6fig:hardware_navigation_trajectories}, the Path Filter Planner and the Aware Planner cover longer distances than the Agnostic Planner. For the Path Filter Planner, the difference is $0.23\,\mathrm{m}$ per iteration and for the Aware Planner even $0.63\,\mathrm{m}$. However, while the Path Filter Planner requires more time for an iteration than the Agnostic Planner, the Aware Planner is faster, namely by $0.4\,\mathrm{s}$. Regarding navigation quality in terms of the deviation from the global path, the Agnostic and the Path Filter Planner are very close to each other, with a maximum RMSE difference of $4\%$.

\begin{table}[b]
	\centering
	\caption{Empirical results. The first five rows refer to the full navigation. Hairpin B and rotation-on-spot are denoted as \emph{hpB} and \emph{rs}.} \label{6tab:main_table}
	\begin{tabular}{
			|M{2.75 cm}
			|S[table-format=2.2]@{\,$\textcolor{gray}{\pm}$\,}
			S[table-format=1.2]
			|S[table-format=2.2]@{\,$\textcolor{gray}{\pm}$\,}
			S[table-format=1.2]
			|S[table-format=2.2]@{\,$\textcolor{gray}{\pm}$\,}
			S[table-format=1.2]|}
		\hline
		Planner & \multicolumn{2}{c|}{\textit{Agnostic}} & \multicolumn{2}{c|}{\textit{Path filter}} & \multicolumn{2}{c|}{\textit{Aware}} \\
		\cline{1-7}
		Time [s] & 30.73 &\textcolor{gray}{0.06}& 31.08 &\textcolor{gray}{0.1} & 30.33 &\textcolor{gray}{0.03} \\
		\hline
		Distance [m] & 8.28 &\textcolor{gray}{0.03} & 8.51  &\textcolor{gray}{0.02} & 8.91 &\textcolor{gray}{0.03}\\
		\cline{1-7}
		RMSE [m] & 0.19 &\textcolor{gray}{0.01} & 0.17 &\textcolor{gray}{0.01}& 0.21  &\textcolor{gray}{0.05}\\
		\cline{1-7}
		Mean Torque [Nm] & 2.35 &\textcolor{gray}{0.04} & 2.38 &\textcolor{gray}{0.08} & 2.26 &\textcolor{gray}{0.05} \\
		\cline{1-7}
		Max. Torque [Nm] & 5.64 &\textcolor{gray}{0.25} & 6.77  &\textcolor{gray}{0.44} & 4.74 &\textcolor{gray}{0.53} \\
		\cline{1-7}
		Mean Torq. (\emph{hpB}) [Nm] & 2.51 &\textcolor{gray}{0.04} & 2.34 &\textcolor{gray}{0.04} & 1.81 &\textcolor{gray}{0.05} \\
		\cline{1-7}
		Max. Torq. (\emph{hpB}) [Nm] & 5.33 &\textcolor{gray}{0.07} & 4.69  &\textcolor{gray}{0.05} & 4.27 &\textcolor{gray}{0.05} \\
		\cline{1-7}
		Mean Torq. (\emph{rs}) [Nm] & 2.71 &\textcolor{gray}{0.04} & 1.98 &\textcolor{gray}{0.1} & 2.07 &\textcolor{gray}{0.02} \\
		\cline{1-7}
		Max. Torq. (\emph{rs}) [Nm] & 7.71 &\textcolor{gray}{0.38} & 7.79  &\textcolor{gray}{0.13} & 5.21 &\textcolor{gray}{0.11} \\
		\hline
	\end{tabular}
\end{table}

\begin{figure}[t]
	\vspace{2mm} 
	\centering
	\includegraphics[width=1\linewidth]{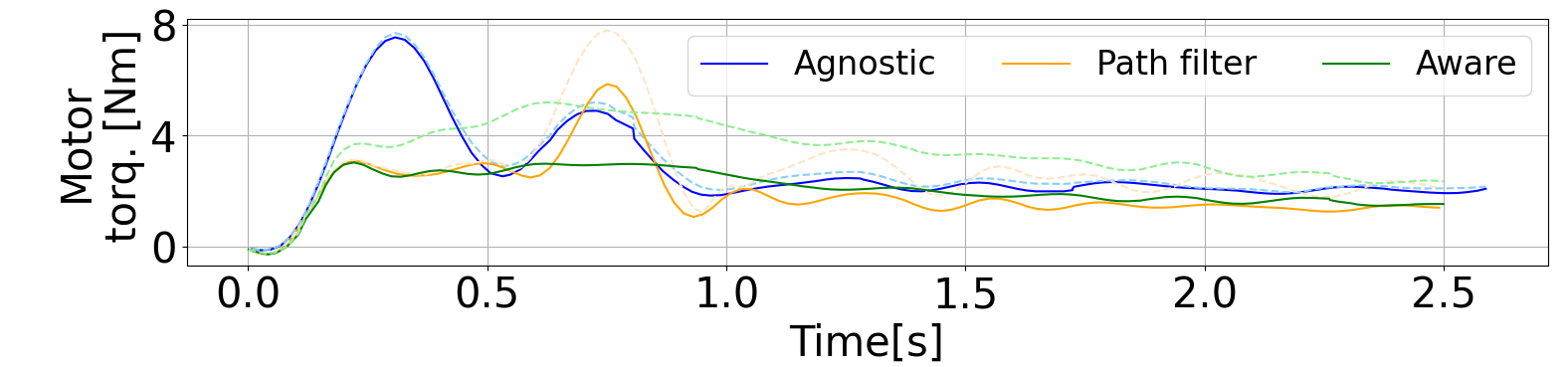}
	\caption{Comparison of averaged-mean (continuous) and -maximum (dashed) torques when rotating on the spot.}
	\label{6fig:hardware_mpc_rotSpot}
\end{figure}

When it comes to motor torques, in comparison to the Agnostic Planner, the Aware Planner reduces averaged-mean and -maximum values by $0.09\,\mathrm{Nm}$ and $0.9\,\mathrm{Nm}$, respectively. These reductions increase up to $0.13\,\mathrm{Nm}$ and $2.03\,\mathrm{Nm}$ if compared to the Path Filter Planner. 

To understand the excessive differences related to the Path Filter Planner, the following phenomena has to be considered: Following the original paper \cite{Schroeder2019_Enhanced_Motion_Control}, the path filter has been implemented for one front caster wheel only, here the left caster wheel. Therefore, in right turns (hairpin A), the inner wheel is not considered by the path filter. As a consequence, the filtered commands violate physical constraints, causing the inbuilt ActiveShuttle's safety controller to interfere. The respective bore torque spike can be visualized around $t=0.5\,\mathrm{s}$ of hairpin A in Fig.~\ref{6fig:torques_navigation}. This phenomenon exemplifies the main disadvantage of using separate, downstream methods compared to the integrated approach of the proposed Aware Planner. 

Seeking a fair motor torque comparison, we have focused on hairpin B. The results are given in the sixth and seventh rows of Table \ref{6tab:main_table} and shown in Fig.~\ref{6fig:torques_navigation}. As expected, the Path Filter Planner requires smaller motor torques than the Agnostic one, and the Aware Planner remains being the one with smallest values for both, averaged-mean and -maximum, cases.

\subsection{Case study II: Rotation on spot}
\noindent Secondly, we focus on the most critical maneuver, i.e., turning on the spot. As depicted in Fig.~\ref{6fig:hardware_mpc_rotSpot} and shown in the last two rows of Table \ref{6tab:main_table}, the Aware and Path Filter Planner require very similar mean torques, approximately a reduction of 26\% with respect to the Agnostic Planner. The lower values are attributed to simultaneous longitudinal and angular motions, instead of rotating while standing still, as the Agnostic Planner does. When it comes to maximum torques, the Aware Planner demands roughly $1.5\,\mathrm{Nm}$ or 30\% less than the other two planners. 

\subsection{Discussion}
\noindent To sum up, we have verified that the Aware Planner achieves, and, in some cases even surpasses, the performance of the Path Filter Planner without significantly degrading the navigation quality. We have also observed an exemplary scenario, where decoupling caster wheel awareness from motion planning, as done in the Path Filter Planner, results in constraint violations. Therefore, the ability to trade off between navigation performance and bore torques, while ensuring the fulfillment of external constraints, accounts for the novelty of the presented approach. 

\section{CONCLUSION}\label{sec:conclusion}
\noindent In this work, we have proposed a method for considering caster wheel reaction torques in motion planning. To do so, a caster-wheel-aware term that accounts for differences between estimated and steady state rolling speeds has been presented. This approach is modular in such a way that it is compatible with MPC-based navigation algorithms and applicable to any mobile robot or car-like vehicle with caster wheels. This is accompanied by a caster wheel rotation angle and rolling speed estimator, enhancing the method's applicability without additional sensors. 

As a proof of concept, an NMPC based trajectory tracking algorithm that includes the caster-wheel-aware term has been formulated. To evaluate the presented method, experiments in a commercial intralogistics SDV have been conducted. Empirical results when comparing this planner, denoted as \textit{Aware Planner}, with a state-of-the-art solution, \textit{Path Filter}, and a baseline caster wheel agnostic planner, \textit{Agnostic planner}, have revealed that: 1) On account of the caster-wheel-aware term and the resulting smoother curvatures, the Aware Planner covers longer distances at higher average speeds and without significantly reducing the navigation quality. 2) When compared to state-of-the-art methods, not only it achieves equivalent performance, but it also embeds the caster wheel awareness in the motion planning stage, and thus, ensures the fulfillment of constraints.

For future work, we plan to replace trajectory tracking with path following, where the predefined path is not parameterized in time. In such a framework, since the motion planner is not biased to be in \textit{a specific location at a given time}, the decision-making power of the caster-wheel-aware term increases. Of great interest is Tunnel Following NMPC \cite{van2019online}, where the trajectory with the lowest reaction torques among the ones inside the corridor could be chosen. This would bring up new scenarios, such as obstacle avoidance. Last but not least, considering that the caster wheel rotation lies in the special Euclidean group SO(2), Lie groups could be used to reformulate the caster-wheel-aware term.

\bibliographystyle{IEEEtran}
\bibliography{IEEEabrv,cw-aware_mpc-planner}

\end{document}